\documentclass[10pt,twocolumn,letterpaper]{article}

\usepackage{cvpr}              

\usepackage[dvipsnames]{xcolor}

\definecolor{cvprblue}{rgb}{0.21,0.49,0.74}
\usepackage[pagebackref,breaklinks,colorlinks,citecolor=cvprblue]{hyperref}

\usepackage{multirow}

\usepackage[accsupp]{axessibility}

\title{VP3D: Unleashing 2D Visual Prompt for Text-to-3D Generation}

\author{Yang Chen, Yingwei Pan, Haibo Yang, Ting Yao, and Tao Mei\\
HiDream.ai Inc.\\
Fudan University\\
{\tt\small \{c1enyang, pandy, tiyao, tmei\}@hidream.ai, yanghaibo.fdu@gmail.com}
}

\begin{document}
\maketitle
\begin{abstract}
Recent innovations on text-to-3D generation have featured Score Distillation Sampling (SDS), which enables the zero-shot learning of implicit 3D models (NeRF) by directly distilling prior knowledge from 2D diffusion models. However, current SDS-based models still struggle with intricate text prompts and commonly result in distorted 3D models with unrealistic textures or cross-view inconsistency issues. In this work, we introduce a novel Visual Prompt-guided text-to-3D diffusion model (VP3D) that explicitly unleashes the visual appearance knowledge in 2D visual prompt to boost text-to-3D generation. Instead of solely supervising SDS with text prompt, VP3D first capitalizes on 2D diffusion model to generate a high-quality image from input text, which subsequently acts as visual prompt to strengthen SDS optimization with explicit visual appearance. Meanwhile, we couple the SDS optimization with additional differentiable reward function that encourages rendering images of 3D models to better visually align with 2D visual prompt and semantically match with text prompt. Through extensive experiments, we show that the 2D Visual Prompt in our VP3D significantly eases the learning of visual appearance of 3D models and thus leads to higher visual fidelity with more detailed textures. It is also appealing in view that when replacing the self-generating visual prompt with a given reference image, VP3D is able to trigger a new task of stylized text-to-3D generation. Our project page is available at \href{https://vp3d-cvpr24.github.io}{https://vp3d-cvpr24.github.io}.
\end{abstract}    
\section{Introduction}
\label{sec:intro}

\begin{figure}[tp]
\begin{center}
\vspace{-0.1in}
\includegraphics[width=0.98\linewidth]{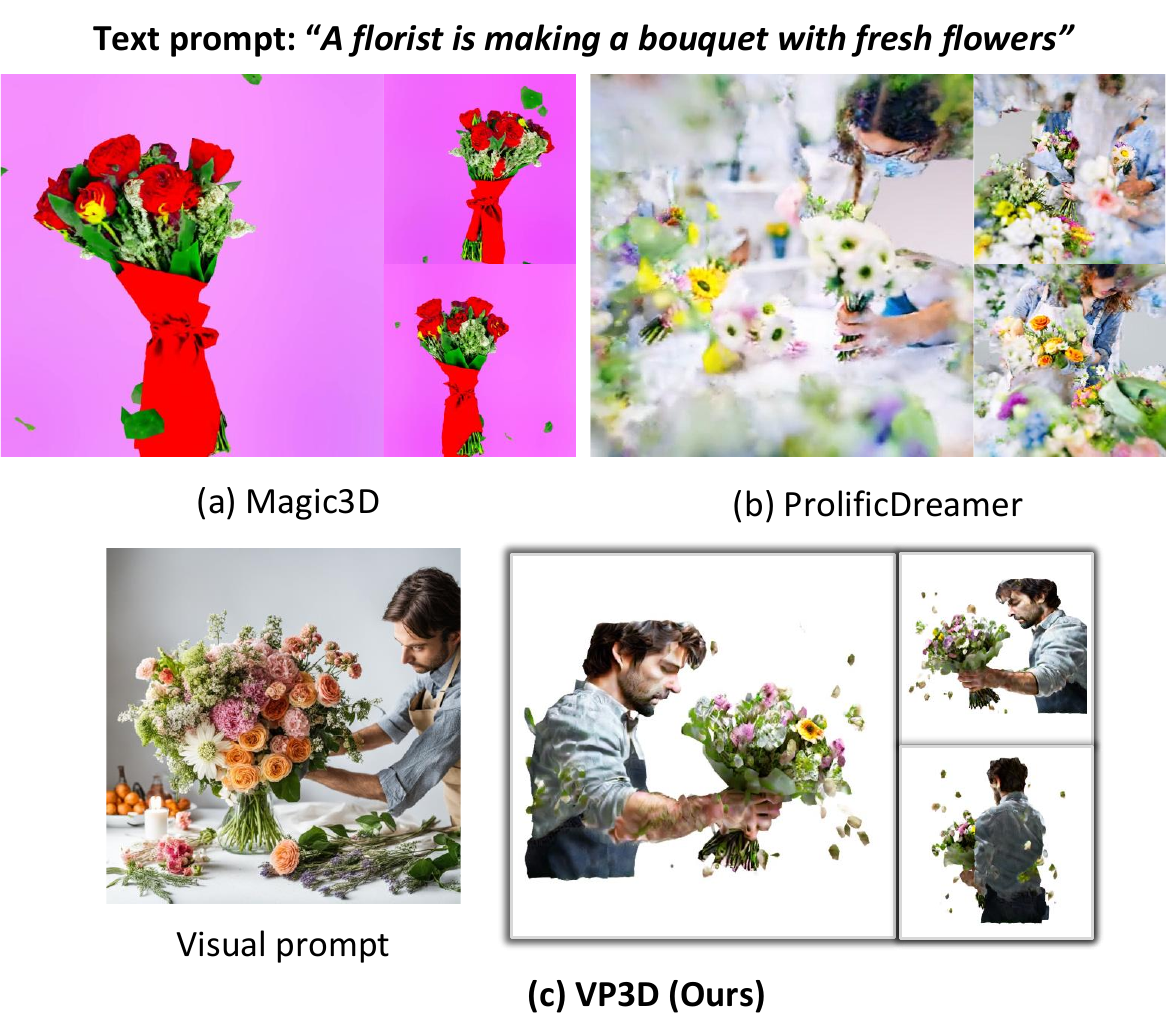}
\vspace{-0.1in}
\end{center}
\vspace{-0.1in}
  \caption{Exisiting text-to-3D generation techniques (e.g., Magic3D \cite{lin2022magic3d} and ProlificDreamer \cite{wang2023prolificdreamer}) often suffer from degenerated results (e.g., over-saturated appearances or inaccurate geometries). Our VP3D novelly integrates a visual prompt to strength score distillation sampling, leading to better 3D results.}
\label{fig:intro}
\vspace{-0.2in}
\end{figure}

Generative Artificial Intelligence (especially for vision content generation) has aroused great attention in computer vision field \cite{chen2019mocycle, chen2019animating,pan2017create,luo2023semantic}, leading to impressive advancements in text-to-image \cite{ramesh2022hierarchical, saharia2022photorealistic, rombach2022high} and text-to-video generation \cite{singer2022make, ho2022imagen, he2022latent}. These accomplishments can be attributed to the availability of large-scale image-text and video-text pair data \cite{bain2021frozen, schuhmann2022laion} and the emergence of robust diffusion-based generative models \cite{sohl2015deep, ho2020denoising, nichol2021improved, ho2022classifier}.
Recently, researchers have gone beyond text-driven image/video generation, and begun exploring diffusion models for text-driven content creation of 3D assets (e.g., text-to-3D generation). This direction paves a new way for practical 3D content creation and has a great potential impact for numerous applications like virtual reality, gaming and Metaverse. Compared to image generation, text-to-3D generation, however, is more challenging, due to the complexities associated with intricate 3D geometry and appearance (i.e., object shapes and textures). Moreover, the collection and annotation of 3D data are somewhat resourcefully expensive and thus cannot be easily scaled up to billion level as image data.

To tackle this issue, a pioneering text-to-3D work (DreamFusion \cite{poole2022dreamfusion}) presents the first attempt of exploiting an off-the-shelf text-to-image diffusion model to generate promising 3D assets in a zero-shot fashion. The key design behind such success is Score Distillation Sampling (SDS), which directly optimizes the implicit 3D model of Neural Radiance Field (NeRF) with prior knowledge distilled from 2D diffusion model. Nevertheless, such distilled prior knowledge is merely driven by the input text prompt, and it is non-trivial to learn high-quality NeRF with distilled SDS supervision. Although several subsequent works \cite{wang2022score, metzer2022latent, lin2022magic3d, chen2023fantasia3d, wang2023prolificdreamer} further upgrade SDS, this kind of SDS-based solution still results in degenerated 3D models with unrealistic/less-detailed textures, especially when feeding intricate text prompts (as seen in Figure \ref{fig:intro} (a-b)).

In this work, we propose to mitigate this limitation through a unique design of visual prompt-guided text-to-3D diffusion model, namely VP3D. Intuitively, ``a picture is worth a thousand words.'' That is, a single image can convey human intentions of visual content creation (e.g., the visual appearance or semantic structure) more effectively than textual sentences. This motivates us to introduce additional guidance of visual prompt, and thus decompose the typical single-shot text-to-3D process into two cascaded processes: first text-to-image generation, and then (text plus image)-to-3D generation. In particular, VP3D first leverages off-the-shelf text-to-image diffusion models to produce a high-fidelity image that reflects extremely realistic appearance with rich details. In the latter process, this synthetic image further acts as 2D visual prompt to supervise SDS optimization of NeRF, coupled with the input text prompt. At the same time, a differentiable reward function is additionally utilized to encourage the rendering images of NeRF to be better aligned with 2D visual prompt (visual appearance consistency) and text prompt (semantic consistency). As illustrated in Figure \ref{fig:intro} (c), we show that the novel visual prompt-guided diffusion process in VP3D significantly enhances the visual fidelity of 3D assets with realistic and rich texture details. Meanwhile, when easing the learning of visual appearance of 3D assets via visual prompt guidance, the optimization of NeRF will focus more on the modeling of geometry, leading to better 3D sharps with cross-view consistency. We believe that the ability of unleashing high-quality visual knowledge in 2D visual prompt is potentially a new paradigm of text-to-3D generation.

As a by-product, we also demonstrate that our VP3D can be readily adapted for a new task of stylized text-to-3D generation. Intriguingly, we simply replace the self-generating image in VP3D with a user-given reference image, and treat it as a new visual prompt to trigger (text plus image)-to-3D generation. In this way, our VP3D is able to produce a stylized 3D asset, which not only semantically aligns with text prompt but also shares similar geometric \& visual style as the reference image.
\section{Related Work}

\textbf{Text-to-3D generation.} Significant advancements have been witnessed in text-to-image generation with 2D diffusion models in recent years \cite{sohl2015deep, ho2020denoising, nichol2021improved, ho2022classifier, ramesh2022hierarchical, saharia2022photorealistic, rombach2022high,chen2023controlstyle}. However, extending these capabilities to 3D content generation poses a substantial challenge, primarily due to the absence of large-scale paired text-3D datasets. To mitigate the reliance on extensive training data, recent works try to accomplish zero-shot text-to-3D generation \cite{poole2022dreamfusion, wang2022score, metzer2022latent, lin2022magic3d,chen2023control3d, chen20233d, yang20233dstyle, chen2023fantasia3d, wang2023prolificdreamer}. Specifically, the pioneering work DreamFusion \cite{poole2022dreamfusion} showcased remarkable achievements in text-to-3D generation through pre-trained text-to-image diffusion models. SJC \cite{wang2022score} concurrently addressed the out-of-distribution problem in lifting 2D diffusion models to perform text-to-3D generation. Following these, several subsequent works have strived to enhance text-to-3D generation further. For instance, Latent-NeRF \cite{metzer2022latent} proposed to incorporate a sketch shape to guide the 3D generation directly in the latent space of a latent diffusion model. Magic3D \cite{lin2022magic3d} presented a coarse-to-fine strategy that leverages both low- and high-resolution diffusion priors to learn the underlying 3D representation. Control3D \cite{chen2023control3d} proposed to enhance user controllability in text-to-3D generation by incorporating additional hand-drawn sketch conditions. ProlificDreamer \cite{wang2023prolificdreamer} presented a principled particle-based variational framework to improve the generation quality. 

Unlike previous works, we formulate the text-to-3D generation process from a new perspective. We first leverage the off-the-shelf text-to-image diffusion models to generate a high-quality image that faithfully matches the input text prompt. This synthetic reference image then serves as a complementary input alongside the text, synergistically guiding the 3D learning process. Moreover, we showcase the remarkable versatility of this novel architecture by effortlessly extending its capabilities to the realm of stylized text-to-3D generation. The resulting 3D asset not only exhibits semantic alignment with the provided text prompt but also masterfully captures the visual style of the reference image. This capability marks another pivotal distinction between our VP3D and previous text-to-3D approaches.

\noindent \textbf{Image-to-3D generation.} Recently, prior works RealFusion \cite{melas2023realfusion}, NeuralLift-360 \cite{xu2023neurallift} and NeRDi \cite{deng2023nerdi} leverage 2D diffusion models to achieve image-to-3D generation. The following work Make-IT-3D \cite{tang2023make} proposed a two-stage optimization framework to improve the generation quality further. Zero-1-to-3 \cite{liu2023zero} finetuned the Stable Diffusion model to enable generating novel views of the input image. It can then be used as a 3D prior model to achieve high-quality image-to-3D generation. Inspired by this, Magic123 \cite{qian2023magic123} proposed to use 2D and 3D priors simultaneously to generate faithful 3D content from the given image. One-2-3-45 \cite{liu2023one} integrated Zero-1-to-3 and a multi-view reconstruction model to accelerate the 3D generation process. It is worth noting that our work is not targeting image-to-3D generation. We utilize a reference image to guide the text-to-3D learning process, instead of directly turning the reference image into 3D content.

\section{VP3D}
\begin{figure*}
    \centering
    \includegraphics[width=0.9\linewidth]{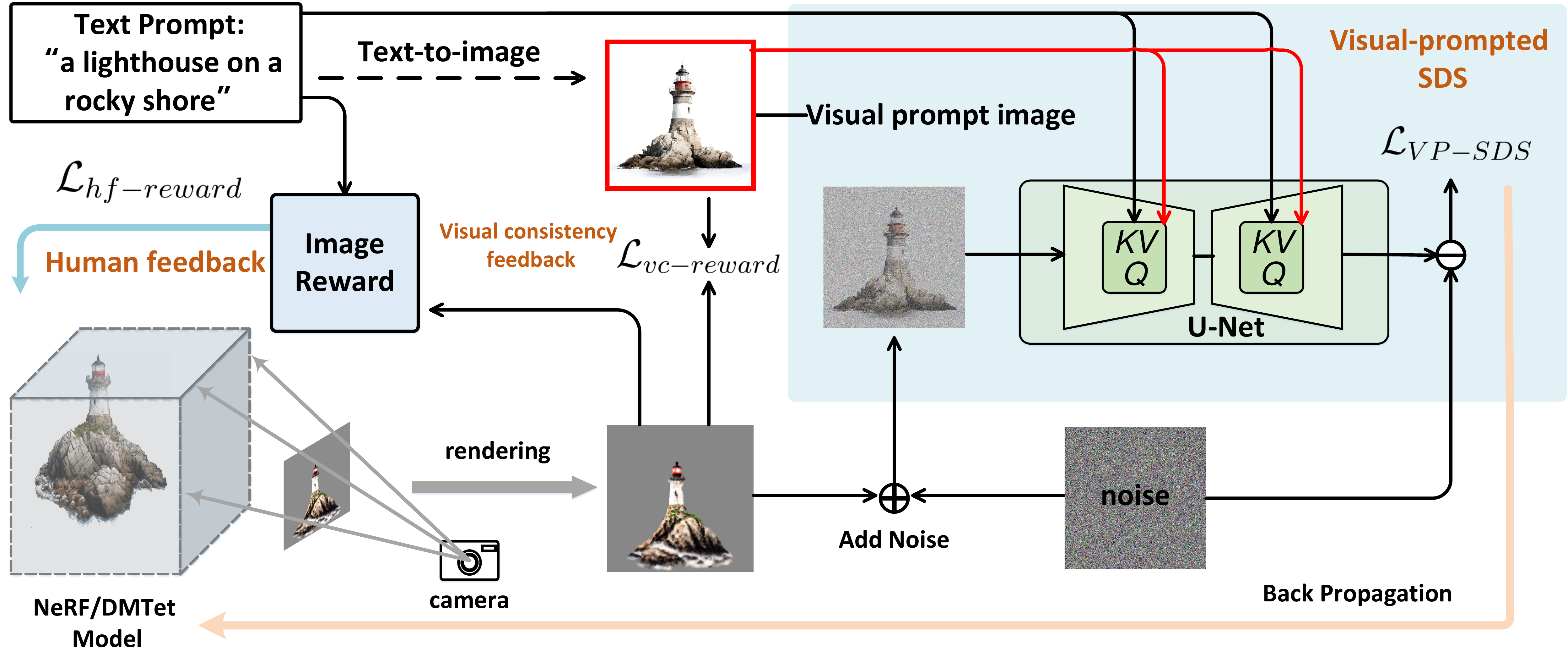}
    \vspace{-8pt}
    \caption{An overview of the proposed VP3D framework for visual prompted text-to-3D generation.}
    \vspace{-0.2in}
    \label{fig:framework}
\end{figure*}

In this section, we elaborate the architecture of our VP3D, which introduces a novel visual prompt-guided text-to-3D diffusion model. 
An overview of our VP3D architecture is depicted in Figure \ref{fig:framework}.
\subsection{Background}

\textbf{Text-to-Image Diffusion Models.} Diffusion models are a family of generative models that are trained to gradually transform Gaussian noise into samples from a target distribution \cite{ho2020denoising}. Given a target data distribution $q(\mathbf{x})$, a \emph{forward diffusion process} is defined to progressively add a small amount of Gaussian noise to the data $\mathbf{x}_0$ sampled from $q(\mathbf{x})$. This process follows a Markov chain $q(\mathbf{x}_{1:T}) = \prod^{T}_{t=1}q(\mathbf{x}_t|\mathbf{x}_{t-1})$ and produces a sequence of latent variables $\mathbf{x}_1, \dots, \mathbf{x}_T$ after $T$ time steps. The marginal distribution of latent variables at time step $t$ is given by $q(\mathbf{x}_t|\mathbf{x}) = \mathcal{N}(\mathbf{x}_t;\alpha_t\mathbf{x}, \sigma^2_t\mathbf{I})$. Thus the noisy sample $\mathbf{x}_t$ can be directly generated through the equation $\mathbf{x}_t=\alpha_t\mathbf{x} + \sigma^2_t\epsilon$, where $\epsilon \sim \mathcal{N}({\bf{0,I}})$, $\alpha_t$ and $\sigma_t$ are chosen parameters such that $\alpha^2_t + \sigma^2_t=1$. After $T$ noise adding steps, $\mathbf{x}_T$ is equivalent to an isotropic Gaussian distribution. Then, a \emph{reverse generative process} is defined to gradually ``denoise'' $X_T$ to reconstruct the original sample. This can be described by a Markov process $p_\phi(\mathbf{x}_{0:T})=p(\mathbf{x}_T)\prod^T_{t=1}p_\phi(\mathbf{x}_{t-1} | \mathbf{x}_t)$, with the conditional probability $p_\phi(\mathbf{x}_{t-1} \vert \mathbf{x}_t) = \mathcal{N}(\mathbf{x}_{t-1}; 
\boldsymbol{\mu}_\phi(\mathbf{x}_t, t), \boldsymbol{\Sigma}_\phi(\mathbf{x}_t, t))$. Commonly, a UNet neural network $\boldsymbol{\epsilon}_\phi(\mathbf{x}_t; t)$ with parameters $\phi$ is used to predict the noise that was used to produce $\mathbf{x}_t$ at time step $t$. Text-to-image diffusion models build upon the above theory to condition the diffusion process with a given text prompt $y$ using classifier-free guidance (CFG) \cite{ho2022classifier}. The corresponding noise predictor is remodeled by:
\begin{equation}
\label{eq:cfg}
\hat{\boldsymbol{\epsilon}}_\phi(\mathbf{x}_t; \mathbf{z}_y, t) = \boldsymbol{\epsilon}_\phi(\mathbf{x}_t; t, \emptyset) + s * (\boldsymbol{\epsilon}_\phi(\mathbf{x}_t; t, \mathbf{z}_y) - \boldsymbol{\epsilon}_\phi(\mathbf{x}_t; t, \emptyset)),
\end{equation}
where $s$ is a scale that denotes the classifier-free guidance weight, $\mathbf{z}_y$ is the corresponding text embedding of the text prompt $y$ and $\emptyset$ indicates the noise prediction without conditioning. The diffusion model $\boldsymbol{\epsilon}_\phi$ is typically optimized by a simplified variant of the variational lower bound of the log data likelihood, which is a Mean Squared Error criterion: 
\begin{equation}
  \mathcal{L}_\mathrm{diff}(\phi) = \mathbb{E}_{\mathbf{x},t,\epsilon}\Bigl[w(t)\|\hat{\boldsymbol{\epsilon}}_\phi(\mathbf{x}_t; y,t) - \epsilon \|^2_2 \Bigr],
\label{equation:diffloss}
\end{equation}
where $w(t)$ is a weighting function that depends on the timestep $t \sim \mathcal{U}(0,1)$ and $\epsilon \sim \mathcal{N}({\bf{0,I}})$.

\noindent \textbf{Score Distillation Sampling.}
A recent pioneering work called DreamFusion \cite{poole2022dreamfusion} introduces Score Distillation Sampling (SDS) that enables leveraging the priors of pre-trained text-to-image diffusion models to facilitate text-to-3D generation. Specifically, let $\theta$ be the learnable parameters of a 3D model (e.g., NeRF) and $g$ be a differentiable rendering function that can render an image $\mathbf{x}=g(\theta; \mathbf{c})$ from the 3D model $\theta$ at a camera viewpoint $\mathbf{c}$. SDS introduces a loss function $\mathcal{L}_{SDS}$ to optimize the parameters $\theta$. Its gradient is defined as follows:
\begin{equation}
\label{eq:sds}
\nabla_{\theta}\mathcal{L}_{SDS} = \mathbb{E}_{t,\epsilon}[ w(t)( \hat{\boldsymbol{\epsilon}}_\phi(\mathbf{x}_t;t,\mathbf{z}_y)-\epsilon)\frac{\partial \mathbf{x}}{\partial \theta}], 
\end{equation}
where $\mathbf{x}_t$ is obtained by perturbing the rendered image $\mathbf{x}$ with a Gaussian noise $\epsilon$ corresponding to the $t$-th timestep of the \emph{forward diffusion process}, $\mathbf{z}_y$ is the conditioned text embedding of given text prompt $y$. Intuitively, the SDS loss estimates an update direction in which the noised version of rendered image $\mathbf{x}$ should be moved towards a denser region in the distribution of real images (aligned with the conditional text prompt y). By randomly sampling views and backpropagating the gradient in Eq. \ref{eq:sds} to the parameters $\theta$ through the differential parametric function $g$, this approach eventually results in a 3D model that resembles the text.

\subsection{Visual-prompted Score Distillation Sampling}

\textbf{Visual Prompt Generation.}
As aforementioned, score distillation sampling plays a key role in text-to-3D generation. Nevertheless, empirical observations \cite{poole2022dreamfusion, wang2023prolificdreamer, he2023t} reveal that SDS still results in degenerated 3D models especially when feeding intricate text prompts. First, SDS-generated results often suffer from over-saturation issues. These issues are, in part, attributed to the necessity of employing a large CFG value (i.e., 100) within the SDS framework \cite{poole2022dreamfusion, wang2023prolificdreamer}. A Large CFG value narrows down the score distillation space to more text-relevant areas. This can mitigate the divergence of diffusion priors in the optimization process, thereby fostering enhanced stability in 3D representation learning. However, this is at the cost of less realistic and diversity generation results, as large CFG values are known to yield over-saturated results \cite{wang2023prolificdreamer}. Second, results generated by SDS still face the risk of text-3D misalignment, such as missing key elements in the scene, especially when text prompts contain multiple objects with specific attributes. 

A fundamental reason behind the aforementioned issues may lie in the substantial distribution gap between text and 3D modalities. Thus it is non-trivial to directly learn
a meaningful 3D scene solely based on a single text prompt. This insight motivates us to introduce an additional visual prompt as a bridge to explicitly establish a connection between the text input and the desired 3D output. Particularly, we leverage off-the-shelf text-to-image diffusion models (e.g., Stable Diffusion) to produce a high-fidelity image that faithfully matches the input text prompt and has an extremely realistic appearance. This image is then used as a visual prompt in conjunction with the input text prompt to jointly supervise the 3D generation process.

\noindent \textbf{Score Distillation Sampling with Visual Prompt.} We now present visual-prompted score distillation sampling that distillation knowledge from a pre-trained diffusion model to optimize a 3D model by considering inputs not only from a text prompt $y$ but also from a visual prompt $v$. To be clear, we restructure the standard SDS-based text-to-3D pipeline by utilizing an image-conditioned diffusion model \cite{ye2023ip} to trigger visual prompt-guided text-to-3D generation. Technically, the visual prompt is first converted to a global image embedding $\mathbf{z}_v$ by the CLIP image encoder \cite{radford2021learning} and a following projection network. This image embedding represents the rich content and style of the visual prompt and has the same dimension as the text embedding $\mathbf{z}_y$ used in the pre-trained text-to-image diffusion model (Stable Diffusion). Following SDS, we first add noise $\epsilon$ to the rendered image of the underlying 3D model according to the random sampled time step $t$ to get a noised image $\mathbf{x}_t$. Then $\mathbf{x}_t$ is input to the diffusion model along with the conditional visual prompt embedding $\mathbf{z}_v$ and text prompt embedding $\mathbf{z}_y$ to estimate the added noise as follows:
\begin{equation}
\small
\begin{aligned}
\tilde{\boldsymbol{\epsilon}}_\phi(\mathbf{x}_t; t, \mathbf{z}_y, \mathbf{z}_v) &= \boldsymbol{\epsilon}_\phi(\mathbf{x}_t; t, \emptyset, \emptyset)\\ 
&+ s * (\boldsymbol{\epsilon}_\phi(\mathbf{x}_t; t, \mathbf{z}_y, \lambda * \mathbf{z}_v)) - \boldsymbol{\epsilon}_\phi(\mathbf{x}_t; t, \emptyset, \emptyset)),
\end{aligned}
\end{equation}
where $s$ is the classifier-free guidance weight, $\lambda \in [0, 1]$ is the visual prompt condition weight, $\phi$ is the parameter of the pre-trained noise predictor $\boldsymbol{\epsilon}_\phi$ and $\boldsymbol{\epsilon}_\phi(\mathbf{x}_t; t, \emptyset, \emptyset)$ denotes the noise prediction without conditioning. 
In this way, our proposed method explicitly incorporates the visual prompt and text prompt in a unified fashion for text-to-3D generation. Consequently, the final gradient of our introduced visual-prompted score distillation sampling (VP-SDS) loss $\theta$ is expressed as:
\begin{equation}
\label{eq:vp-sds}
\nabla_{\theta}\mathcal{L}_{VP-SDS} = \mathbb{E}_{t,\epsilon}[ w(t)(\tilde{\boldsymbol{\epsilon}}_\phi(\mathbf{x}_t;t,\mathbf{z}_y,\mathbf{z}_v)- \boldsymbol{\epsilon})\frac{\partial \mathbf{x}}{\partial \theta}],
\end{equation}
where $w(t)$ is a scheduling coefficient. 

\noindent \textbf{Comparison with SDS.} Comparing the update gradient of SDS (Eq. \ref{eq:sds}) and VP-SDS (Eq. \ref{eq:vp-sds}), SDS is a special case of our VP-SDS by setting $\lambda = 0$ where the visual prompt condition is neglected. In accordance with the theoretical analysis presented in \cite{poole2022dreamfusion, wang2023prolificdreamer}, the mode-seeking nature of SDS necessitates a large CFG to ensure that the pre-trained diffusion model $\boldsymbol{\epsilon}_\phi$ delivers a ``sharp'' updating direction for the underlying 3D model. Nevertheless, a large CFG, in turn, results in poor-quality samples and thus a ``degraded'' update direction. 
In contrast, VP-SDS leverages the additional visual prompt to narrow down the distillation space of $\boldsymbol{\epsilon}_\phi$ into a more compact region that aligns tightly with the visual prompt. Meanwhile, the distillation space is also refined by the visual prompt as it reflects realistic appearances with rich details. Therefore, the updating direction derived from our VP-SDS is not only ``sharp'' but also ``fine'', which can obtain much better 3D generation results than SDS. 

Notably, a recent work ProlificDreamer \cite{wang2023prolificdreamer} presents variational score distillation (VSD) to address the aforementioned issues in SDS. However, VSD needs to train an additional diffusion model using LoRA \cite{hu2106low} during the optimization process, which incurs a considerable computational overhead compared to SDS. Instead, the additional computational cost of our VP-SDS is nearly negligible, making it computationally more efficient than VSD.   

\noindent \textbf{View-dependent Visual Prompting.} Apart from the over-saturation problem discussed above, existing text-to-3D methods are known to also suffer from the multi-view inconsistency problem (e.g., the multi-face Janus problem). This arises from the fact that the underlying prior diffusion model is exclusively trained on individual 2D images and therefore lacks 3D awareness. To alleviate this issue, existing text-to-3D methods \cite{poole2022dreamfusion, wang2022score, lin2022magic3d, wang2023prolificdreamer} always employ diffusion loss with view-dependent text conditioning, which is to append ``front view'', ``side view'', or ``back view'' to the input text based on the location of the randomly sampled camera. Inspired by this, we devise a view-dependent visual prompting strategy to further mitigate the view inconsistency problem in collaboration with our introduced VP-SDS. Technically, given the input visual prompt (assuming it is shot from the front view), we use a view-conditioned 2D diffusion model, Zero-1-to-3 \cite{liu2023zero}, to transform it into left-side, right-side and backward views. Then we fed different visual prompts into VP-SDS (Eq. \ref{eq:vp-sds}) depending on the corresponding sampled camera viewpoint. For instance, when the azimuth angle $\gamma_{cam} \in [0^\circ, 360^\circ]$ of the camera position falls in the range near $180^\circ$ ($0^\circ$ denotes the front view), we feed the generated back view counterpart of the input visual prompt into Eq \ref{eq:vp-sds}. In this way, the inherent 3D geometry information contained in the multi-view visual prompts is encoded into the 3D representation learning through view-dependent VP-SDS, leading to better view consistency in the 3D generation.

\subsection{Learning with Reward Feedback}
To further encourage rendered images of the underlying 3D model that are high fidelity and well aligned with the input visual prompt and text prompt, we devise two types of differentiable reward functions that complement the aforementioned VP-SDS objective.

\noindent \textbf{Human Feedback Reward.}
Recent practice has shown the capability of improving text-to-image models with human feedback \cite{xu2023imagereward}. Particularly, it first trains a \emph{reward model} on a large dataset comprised of human assessments of text-image pairs. Such a reward model thus has the ability to measure the quality of the generated samples in terms of both image fidelity and image-text alignment. Consequently, it can then be used to fine-tune diffusion models to maximize the predicted scores of the reward model through differentiable reward functions, leading to better generation results. Motivated by this, we go one step further to utilize the open-sourced reward model $\mathbf{r}$ in ImageReward \cite{xu2023imagereward} for text-to-3D generation. Specifically, we introduce a human feedback reward loss as follows:
\begin{equation}
\label{eq:refl}
\mathcal{L}_{hf-reward} = \mathbb{E}_{\mathbf{c}}[ \psi(\mathbf{r}(\mathbf{x}, y))],
\end{equation}
where $\mathbf{x} = g(\theta; \mathbf{c})$ is a rendered image by the underlying 3D model $\theta$ from an arbitrary viewpoint $\mathbf{c}$, $y$ is the conditional text prompt and $\psi$ is a differentiable reward-to-loss map function as in \cite{xu2023imagereward}. Intuitively, minimizing the loss in Eq. \ref{eq:refl} encourages the rendered image $\mathbf{x}$ to get a higher reward score from the reward model $\mathbf{r}$, which means the underlying 3D model should update towards the refined direction where the renderings have high appearance fidelity and faithfully match the input text prompt.

\noindent \textbf{Visual Consistency Reward.} 
Given that the above human feedback reward only takes into account the input text prompt, we further devised a visual consistency reward to fully leverage the visual prompt as well, since text prompts cannot capture all appearance details. Technically, we adopt a pre-trained self-supervised vision transformer DINO-ViT \cite{caron2021emerging} to extract the visual features $F_{dino}(v)$ and $F_{dino}(\mathbf{x})$ of the input visual prompt $v$ and rendered image $\mathbf{x}$, respectively. Then we penalize the feature-wise difference between them at the visual prompt viewpoint: 
\begin{equation}
\label{eq:visual_consistency}
\mathcal{L}_{vc-reward} = ||F_{dino}(\mathbf{x})-F_{dino}(\mathbf{v})||^2.
\end{equation}
By imposing such visual consistency loss, we encourage the underlying 3D model to adhere to the plausible shape and appearance properties conveyed by the visual prompt. 

\subsection{3D Representation and Training}
Inspired by \cite{lin2022magic3d}, we adopt a two-stage coarse-to-fine framework for text-to-3D generation with two different 3D scene representations. At the coarse stage, we leverage Instant-NGP \cite{muller2022instant} as 3D representation, which is much faster to optimize compared to the vanilla NeRF \cite{mildenhall2020nerf} and can recover complex geometry. In the fine stage, we leverage DMTet as the 3D representation to further optimize a high-fidelity mesh and texture. Specifically, the 3D shape and texture represented in DMTet are first initialized from the density field and color field of the coarse stage, respectively \cite{lin2022magic3d}.

During the optimization process in each stage,  we first render images from the underlying 3D model through differentiable rasterizers at arbitrary camera poses and optimize the 3D model with a combination of losses:  
\begin{equation}
\label{eq:loss_fine}
\mathcal{L}_{fine} =\mathcal{L}_{VP-SDS} + \lambda_1 \mathcal{L}_{vc-reward} + \lambda_2 \mathcal{L}_{hf-reward},
\end{equation}
where $\lambda_1, \lambda_2$ are the trade-off parameters.

\begin{table*}[h]
    \centering  
    \caption{The quantitative results of our method and baselines on T$^3$Bench~\cite{he2023t}.}
    \vspace{-0.1in}
    \resizebox{\linewidth}{!}{
        \begin{tabular}{l|ccc|ccc|ccc}
            \toprule[2pt]
            \multirow{2}{*}{Method} & \multicolumn{3}{c|}{\emph{Single Object}} & \multicolumn{3}{c|}{\emph{Single Object with Surroundings}} & \multicolumn{3}{c}{\emph{Multiple Objects}}  \\
             & Quality & Alignment & Average  & Quality & Alignment & Average  & Quality & Alignment & Average   \\ \midrule
            DreamFusion~\cite{poole2022dreamfusion} & 24.9 & 24.0 & 24.4 & 19.3 & 29.8 & 24.6 & 17.3 & 14.8 & 16.1  \\
            SJC~\cite{wang2022score} & 26.3 & 23.0 & 24.7 & 17.3 & 22.3 & 19.8 & 17.7 & 5.8  & 11.7  \\
            LatentNeRF~\cite{metzer2022latent} & 34.2 & 32.0 & 33.1 & 23.7 & 37.5 & 30.6 & 21.7 & 19.5 & 20.6  \\
            Fantasia3D~\cite{chen2023fantasia3d} & 29.2 & 23.5 & 26.4 & 21.9 & 32.0 & 27.0 & 22.7 & 14.3 & 18.5  \\
            ProlificDreamer~\cite{wang2023prolificdreamer}  & 51.1 & 47.8 & 49.4 & 42.5 & 47.0 & 44.8 & 45.7 & 25.8 & 35.8  \\
            \midrule
            Magic3D~\cite{lin2022magic3d} & 38.7 & 35.3 & 37.0 & 29.8 & 41.0 & 35.4 & 26.6 & 24.8 & 25.7  \\
            \textbf{VP3D (Ours)}  & \textbf{54.8} & \textbf{52.2} &  \textbf{53.5} &  \textbf{45.4} & \textbf{50.8} &  \textbf{48.1} & \textbf{49.1} & \textbf{31.5} & \textbf{40.3} \\
            \bottomrule[2pt]
        \end{tabular}
    }
    \vspace{-0.1in}
    \label{tb:t3bench}
\end{table*}
\section{Experiments}
In this section, we evaluate the effectiveness of our VP3D for text-to-3D generation via extensive empirical evaluations. We first show both quantitative and qualitative results of VP3D in comparison to existing techniques on the newly released text-to-3D benchmark (T$^3$Bench \cite{he2023t}). Next, we conduct ablation studies to validate each design in VP3D. Finally, we demonstrate the extended capability of VP3D for stylized text-to-3D generation.

\subsection{Experimental Settings}
\textbf{Implementation Details.} 
In the coarse and fine stage, the underlying 3D models are both optimized for 5000 iterations using Adam optimizer with 0.001 learning rate. The rendering resolutions are set to $128 \times 128$ and $512 \times 512$ for coarse and fine stage, respectively. We implement the underlying Instant-NGP and DMTet 3D representation mainly based on the Stable-DreamFusion codebase \cite{stable-dreamfusion}. $\lambda_1$ is set to 0.1 in the coarse stage and 0.01 in the fine stage. $\lambda_2$ is linearly increased from 0.001 to 0.01 during the optimization process. The visual prompt condition weight is set to 0.5 in all experiments. 

\noindent \textbf{Evaluation Protocol.}
Existing text-to-3D generation works commonly examine their methods over the CLIP R-Precision score \cite{jain2022zero}, which is an automated metric for the consistency of rendered images with respect to the input text. However, this text-image alignment-based metric cannot faithfully represent the overall 3D quality. For example, CLIP-based text-to-3D methods can also achieve high CLIP R-Precision scores even if the resulting 3D scenes are unrealistic and severely distorted \cite{poole2022dreamfusion}. Taking this into account, we instead conduct experiments on a newly open-sourced benchmark: T$^3$Bench~\cite{he2023t}, which is the first comprehensive text-to-3D benchmark containing 300 diverse text prompts of three categories (single object, single object with surroundings, and multiple objects).

T$^3$Bench provides two automatic metrics (quality and alignment) based on the rendered multi-view images to assess the subjective quality and text alignment. The quality metric utilizes a combination of multi-view text-image scores and regional convolution to effectively identify quality and view inconsistency. The alignment metric employs a 3D captioning model and a Large Language Model (i.e., GPT-4) to access text-3D consistency. Following this, we also leverage the quality and alignment metric to quantitatively compare our VP3D against baseline methods. 

\begin{figure*}[htb]
\begin{center}
\includegraphics[width=0.9\linewidth]{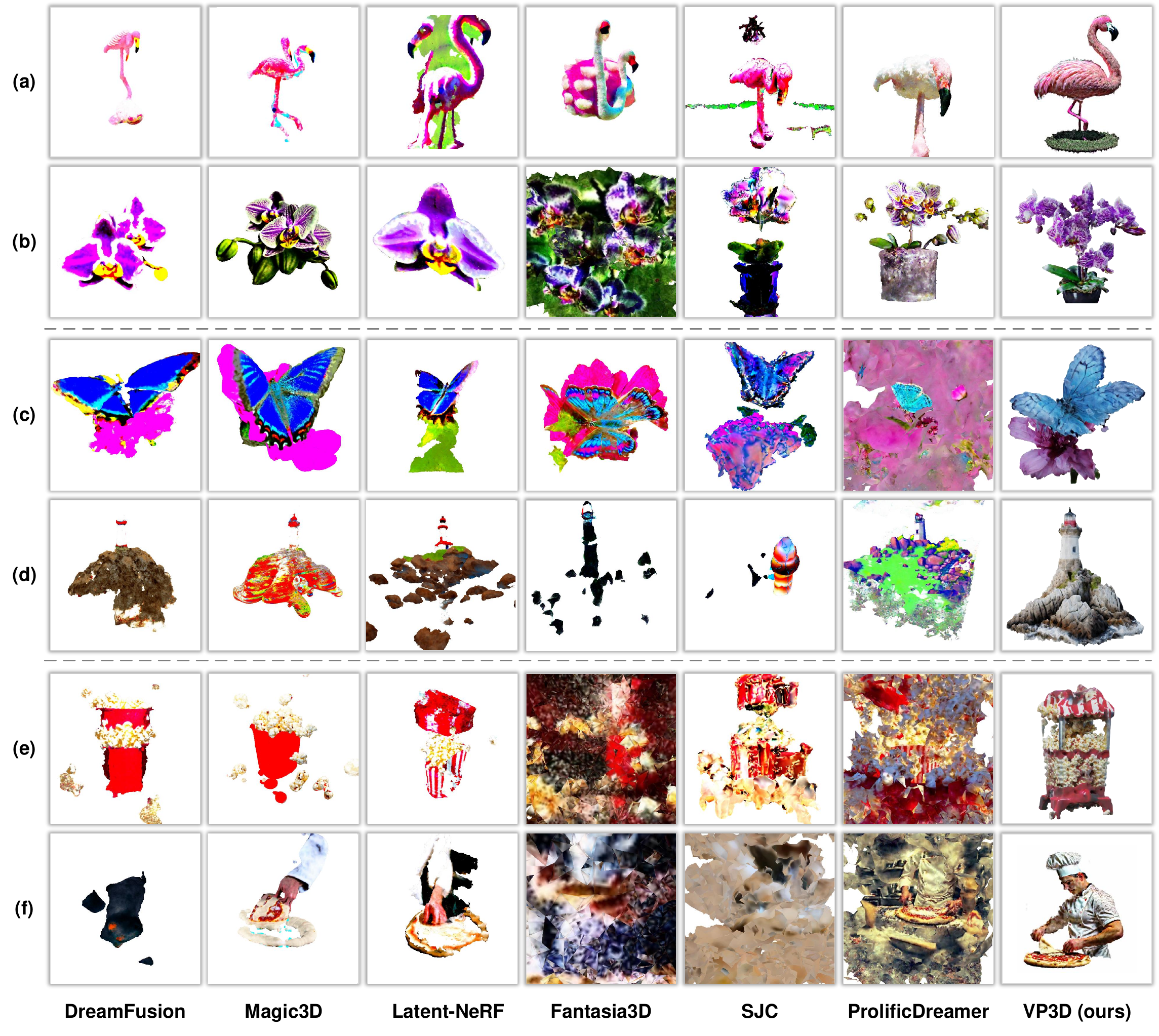}
\end{center}
\vspace{-0.2in}
\caption{Comparisons on qualitative results of our VP3D with other text-to-3D techniques on T$^3$Bench \cite{he2023t}. The prompts are (a) ``A fuzzy pink flamingo lawn ornament'', (b) ``A blooming potted orchid with purple flowers'', (c) ``A blue butterfly on a pink flower'',(d) ``A lighthouse on a rocky shore'',(e) ``Hot popcorn jump out from the red striped popcorn maker'',(f) ``A chef is making pizza dough in the kitchen''. (a-b), (c-d), (e-f) belongs to the \emph{Single Object}, \emph{Single Object with Surr} and \emph{Multi Objects} category in T$^3$Bench, respectively. }
\label{fig:qualitative_comparisons}
\vspace{-0.1in}
\end{figure*}

\noindent \textbf{Baselines.} To evaluate our method, we compare our VP3D with six state-of-the-art text-to-3D generation methods: DreamFusion \cite{poole2022dreamfusion}, SJC \cite{wang2022score}, LatentNeRF \cite{metzer2022latent}, Fantasia3D \cite{chen2023fantasia3d}, Magic3D \cite{lin2022magic3d} and ProlificDreamer \cite{wang2023prolificdreamer}. Specifically, DreamFusion \cite{poole2022dreamfusion} firstly introduces score distillation sampling (SDS) that enables leveraging 2D diffusion model (Imagen \cite{ho2022imagen}) to optimize a NeRF \cite{mildenhall2020nerf}. SJC \cite{wang2022score} concurrently addresses the out-of-distribution problem in SDS and utilizes an open-sourced diffusion model (Stable Diffusion) to optimize a voxel NeRF. Latent-NeRF \cite{metzer2022latent} first brings NeRF to the latent space to harmonize with latent diffusion models, then refines it in pixel space. Magic3D \cite{lin2022magic3d} extends DreamFusion with a coarse-to-fine framework that first optimizes a low-resolution NeRF model and then a high-resolution DMTet model via SDS. Fantasia3D \cite{chen2023fantasia3d} disentangles the SDS-based 3D learning into geometry and appearance learning. ProlificDreamer \cite{wang2023prolificdreamer} upgrades DreamFusion by a variational score distillation (VSD) loss that treats the underlying 3D scene as a random variable instead of a single point as in SDS.

\subsection{Quantitative Results}
The quantitative performance comparisons of different methods for text-to-3D generation are summarized in Table \ref{tb:t3bench}. Overall, our VP3D consistently achieves better performances against existing techniques across all evaluation metrics and prompt categories. Remarkably, VP3D achieves an absolute quality-alignment average score improvement of $4.1\%$, $3.3\%$, and $4.5\%$ against the best competitor ProlificDreamer across the three text prompt categories, respectively, which validates the effectiveness of our overall proposals. More importantly, while VP3D employs the same NeRF $\&$ DMTet 3D representation and coarse-to-fine training scheme as the baseline method Magic3D, it significantly outperforms Magic3D by achieving $53.5\%$, $48.1\%$, and $40.3\%$ average scores, representing a substantial improvement over Magic3D's average scores of $37.0\%$, $35.4\%$, and $25.7\%$. The results generally highlight the key advantage of introducing visual prompts in lifting 2D diffusion models to perform text-to-3D generation.

\begin{figure}
    \begin{center}
    \includegraphics[width=0.47\textwidth]{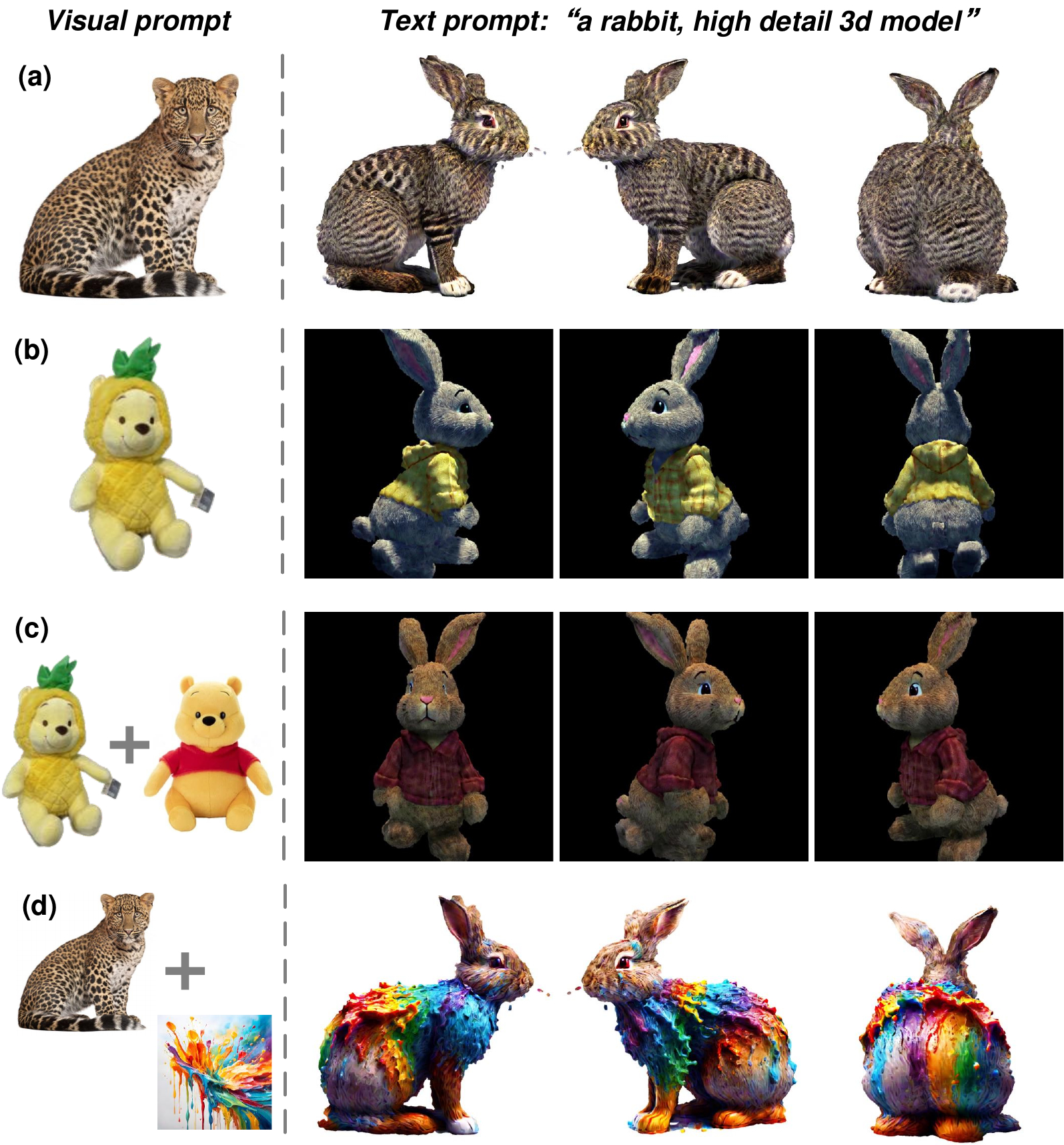}
    \end{center}
\vspace{-0.1in}
           \caption{Stylized text-to-3D generation results of our VP3D.}
           \vspace{-0.1in}
    \label{fig:stylized}
\end{figure}

Specifically, DreamFusion and SJC enable the zero-shot learning of implicit 3D models by distilling prior knowledge from 2D diffusion models. However, the generated 3D scenes have relatively low quality and alignment scores, especially in complex scenarios where the text prompt contains multiple objects or surroundings. Latent-NeRF employs score distillation in the latent space and then back to pixel space to further refine the 3D model, leading to better results. The aforementioned three methods only utilize implicit 3D representations (NeRFs). In contrast, Magic3D adopts textured mesh DMTet as 3D representation for enabling high-resolution optimization and exhibits better performances across all three prompt categories. Fantasia3D also capitalizes on DMTet for geometry learning and then leverages BRDF for appearance learning in a disentangled manner. While Fantasia3D achieves better average scores than DreamFusion and SJC, it fails to create high-fidelity results in complex scenes (e.g., ``multiple objects''). ProlificDreamer further boosts the performance by training an additional diffusion model during the optimization process to realize a principled particle-based variational score distillation loss. However, our VP3D still outperforms ProlificDreamer across all evaluation metrics and prompt sets, which confirms the effectiveness of our VP3D.

\subsection{Qualitative Results}
The qualitative comparisons for text-to-3D generation are presented in Figure~\ref{fig:qualitative_comparisons}. As can be seen, our VP3D generally produces superior 3D scenes with plausible geometry and realistic textures when compared with baseline methods. Specifically, DreamFusion suffers from a severely over-saturated problem and has difficulty generating complex geometry. Magic3D and Latent-NeRF slightly alleviate these issues through their higher-resolution DMTet and pixel space refinement, respectively. While Fantasia3D and SJC can generate richer textures than DreamFusion, the geometric quality of the generated 3D scenes falls short of expectations. Notably, ProlificDreamer trains an additional diffusion model during the optimization process to perform variational score distillation (VSD) instead of SDS, achieving satisfactory single-object objects. However, the use of VSD at times introduces excessive irrelevant information or geometry noise in more complex scenarios. In contrast, we can clearly observe that the generated 3D scenes by VP3D faithfully match the input text prompt with plausible geometry and realistic appearance, which demonstrates the superiority of VP3D over state-of-the-art methods and its ability to generate high-quality 3D content.

\subsection{Ablation Study}
\begin{figure}
    \begin{center}
    \includegraphics[width=0.46\textwidth]{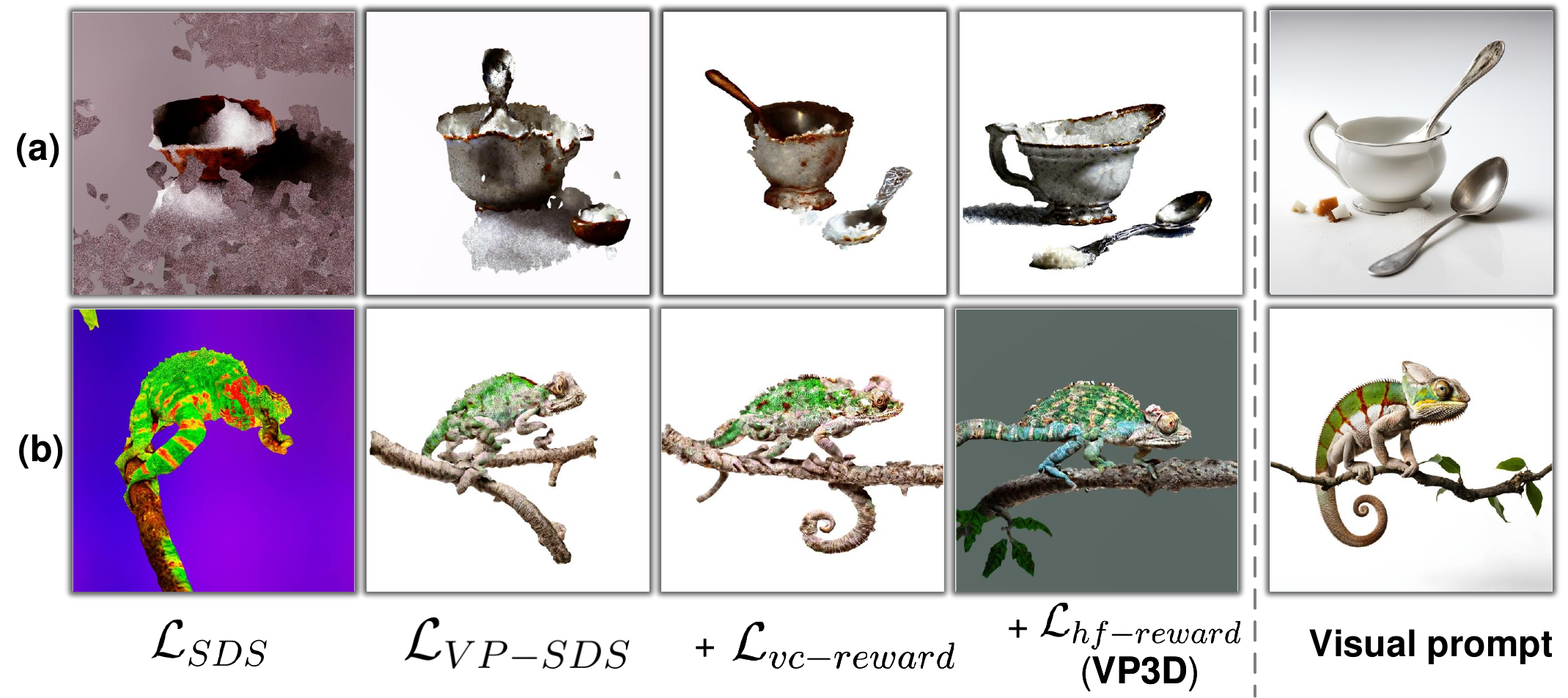}
    \end{center}
\vspace{-0.1in}
           \caption{Comparisons on qualitative results by using different ablated runs of our VP3D. The text prompts are (a) ``A broken tablespoon lies next to an empty sugar bowl'' and (b) ``A chameleon perched on a tree branch''.}
           \vspace{-0.2in}
    \label{fig:ablation}
\end{figure}

Here we investigate how each design in our VP3D influences the overall generation performance. We depict the qualitative results of each ablated run in Figure \ref{fig:ablation}. $\mathcal{L}_{SDS}$ is our baseline model that employs vanilla score distillation sampling loss. As can be seen, the generated 3D scene is over-saturated and geometry unreasonable. Instead, when $\mathcal{L}_{VP-SDS}$ is employed, the generation quality is clearly enhanced in terms of both geometry and appearance. This highlights the critical effectiveness of our proposed visual-prompted score distillation sampling. Nevertheless, the resulting 3D scenes by $\mathcal{L}_{VP-SDS}$ are still not satisfying enough. By utilizing additional visual consistency and human feedback reward functions $\mathcal{L}_{vc-reward}$ (Eq. \ref{eq:visual_consistency}) and $\mathcal{L}_{hf-reward}$ (Eq. \ref{eq:refl}), the generation quality is gradually improved. The results basically validate the effectiveness of these two complementary factors.

\subsection{Extension to Stylized Text-to-3D Generation}
In this section, we demonstrate that another advantage of our VP3D is its remarkable versatility in 3D generation as it can be readily adapted for a new task of stylized text-to-3D generation. The main difference is that the visual prompt is no longer generated from the text prompt but from a user-specified reference image. We also empirically discard the loss in Eq. \ref{eq:refl} to eliminate the strictly text-image alignment constraint. In this way, our VP3D can integrate the visual cues contained in the reference image into text-to-3D generation and produce a stylized 3D asset. This asset not only semantically aligns with the text prompt but also reflects visual and geometry properties in the reference image. 

Figure \ref{fig:stylized} shows our stylized text-to-3D generation results. Our VP3D can generate diverse and stylized 3D assets by giving different visual prompts to the same text prompt. As shown in Figure \ref{fig:stylized} (a-b), the generated result semantically is a rabbit that adheres to the text prompt but also inherits some visual cues of the visual prompt. To be clear, the generated 3D rabbits have somewhat consistent geometry pose and appearance texture with the object in the visual prompt. For example, in Figure \ref{fig:stylized} (b), the generated rabbit mirrors the ``hugging pose'' of the reference image and also has the same style of ``crescent-shaped eyebrows'' and ``yellow plaid jacket'' as in the reference image. In Figure \ref{fig:stylized} (c-d), we showcase the versatility of our VP3D by seamlessly blending styles from different visual prompts. Take Figure \ref{fig:stylized} (d) as an instance, we use the leopard image as a visual prompt in the coarse stage and then replace it with an oil painting image in the fine stage. Our VP3D finally resulted in a 3D rabbit that not only has a consistent pose with the leopard but also a colorful oil painting style texture. The stylized 3D generation ability distinct our VP3D from previous text-to-3D approaches and can lead to more creative and diverse 3D content creation. 

\section{Conclusion}
In this work, we propose VP3D, a new paradigm for text-to-3D generation by leveraging 2D visual prompts. We first capitalize on 2D diffusion models to generate a high-quality image from input text. This image then acts as a visual prompt to strengthen the 3D model learning with our devised visual-prompted score distillation sampling. Meanwhile, we introduce additional human feedback and visual consistency reward functions to encourage the semantic and appearance consistency between the 3D model and input visual $\&$ text prompt. Both qualitative and quantitative comparisons on the T$^3$Bench benchmark demonstrate the superiority of our VP3D over existing SOTA techniques.

{
    \small
    \bibliographystyle{ieeenat_fullname}
    \bibliography{main}
}


\end{document}